\documentclass{article}
\usepackage{spconf,amsmath,graphicx}
\usepackage{bm}
\usepackage{amsfonts}       % blackboard math symbols
\usepackage{nicefrac}       % compact symbols for 1/2, etc.
\usepackage{amsthm}
\usepackage{mathtools}
\usepackage{empheq}
\usepackage{amssymb}
\usepackage{macros}
\usepackage[ruled,vlined]{algorithm2e}
\usepackage{url}
\usepackage{subcaption}
\usepackage{booktabs}
\newcommand{\bx}{\bm{x}}
\newcommand{\bz}{\bm{z}}
\newcommand{\by}{\bm{y}}
\newcommand{\bv}{\bm{v}}
\newcommand{\bX}{\bm{X}}
\newcommand{\bG}{\bm{G}}
\newcommand{\bP}{\bm{P}}
\newcommand{\bW}{\bm{W}}

\newcommand{\lr}{\gamma} % learning rate
\newcommand{\efflr}{\gamma_{\text{eff}}} % effective learning rate
\newcommand{\glr}{\alpha} % slow/global learning rate
\newcommand{\cp}{\tau} % inner loop length
\newcommand{\nworker}{m} % number of worker nodes
\newcommand{\obj}{F}
\newcommand{\tg}{\nabla F}
\newcommand{\sg}{g}
\newcommand{\lip}{L}
\newcommand{\vbnd}{\sigma^2}
\newcommand{\eg}{\textit{e.g.}}
\newcommand{\ie}{\textit{i.e.}}

\newcommand{\alg}{Overlap-Local-SGD}
\newcommand{\spgap}{\zeta}

\usepackage{cleveref}
\Crefname{tab}{Table}{Tables}
\crefname{equation}{}{}
\Crefname{equation}{}{}
\crefname{thm}{theorem}{theorems}
\Crefname{thm}{Theorem}{Theorems}
\crefname{clm}{claim}{claims}
\Crefname{clm}{Claim}{Claims}
\Crefname{coro}{Corollary}{Corollaries}
\Crefname{lem}{Lemma}{Lemmas}
\Crefname{sec}{Section}{Sections}
\crefname{app}{appendix}{appendices}
\Crefname{app}{Appendix}{Appendices}
\Crefname{part}{Part}{Parts}
\crefname{prop}{proposition}{propositions}
\Crefname{prop}{Proposition}{Propositions}
\Crefname{propty}{Property}{Properties}
\crefname{figure}{fig.}{figures}
\Crefname{figure}{Figure}{Figures}
\crefname{defn}{definition}{definitions}
\Crefname{defn}{Definition}{Definitions}
\crefname{fact}{fact}{facts}
\Crefname{fact}{Fact}{Facts}
\crefname{appendix}{appendix}{appendices}
\Crefname{appendix}{Appendix}{Appendices}
\crefname{algo}{algorithm}{algorithms}
\Crefname{algo}{Algorithm}{Algorithms}
\crefname{algorithm}{algorithm}{algorithms}
\Crefname{algorithm}{Algorithm}{Algorithms}
\crefname{conj}{conjecture}{conjectures}
\Crefname{conj}{Conjecture}{Conjectures}
\crefname{obs}{observation}{observations}
\Crefname{obs}{Observation}{Observations}
\crefname{assump}{assumption}{assumptions}
\Crefname{assump}{Assumption}{Assumptions}
\crefname{rem}{remark}{remarks}
\Crefname{rem}{Remark}{Remarks}

\usepackage[colorinlistoftodos,prependcaption]{todonotes}

\title{Overlap Local-SGD: An Algorithmic Approach to Hide Communication Delays in Distributed SGD}

\name{Jianyu Wang \qquad Hao Liang \qquad Gauri Joshi}

\address{Carnegie Mellon University, Pittsburgh, USA \\
	    \{jianyuw1, hliang2, gaurij\}@andrew.cmu.edu}

\begin{document}
\ninept

\maketitle

\begin{abstract}
Distributed stochastic gradient descent (SGD) is essential for scaling the machine learning algorithms to a large number of computing nodes. However, the infrastructures variability such as high communication delay or random node slowdown greatly impedes the performance of distributed SGD algorithm, especially in a wireless system or sensor networks. In this paper, we propose an algorithmic approach named {\alg} (and its momentum variant) to overlap the communication and computation so as to speedup the distributed training procedure. The approach can help to mitigate the straggler effects as well. We achieve this by adding an anchor model on each node. After multiple local updates, locally trained models will be pulled back towards the synchronized anchor model rather than communicating with others. Experimental results of training a deep neural network on CIFAR-10 dataset demonstrate the effectiveness of {\alg}. We also provide a convergence guarantee for the proposed algorithm under non-convex objective functions.

% \textit{A full version of this paper with additional examples and proofs is accessible at: \url{andrew.cmu.edu/user/gaurij/overlap_local_SGD.pdf}}. 
\end{abstract}


\section{Introduction}
Distributed optimization with stochastic gradient descent (SGD) is the backbone of the state-of-the-art supervised learning algorithms, especially when training large neural network models on massive datasets \cite{liu2019roberta,radford2019language}. The widely adopted approach now is to let worker nodes compute stochastic gradients in parallel, and average them using a parameter server \cite{li2014scaling} or a blocking communication protocol \textsc{AllReduce} \cite{goyal2017accurate}. Then, the model parameters are updated using the averaged gradient. This classical parallel implementation is referred as \emph{fully synchronous SGD}. However, in a wireless system where the computing nodes typically have low bandwidth and poor connectivity, the high communication delay and unpredictable nodes slowdown may greatly hinder the benefits of parallel computation \cite{vogels2019powersgd,dutta2018slow,ferdinand2019anytime,amiri2019computation}. It is imperative to make distributed SGD to be fast as well as robust to the system variabilities.

A promising approach to reduce the communication overhead in distributed SGD is to reduce the synchronization frequency among worker nodes. Each node maintains a local copy of the model parameters and performs $\cp$ local updates (only using local data) before synchronizing with others. Thus, in average, the communication time per iteration is directly reduced by $\cp$ times. This method is called \emph{Local SGD} or \emph{periodic averaging SGD} in recent literature \cite{zhou2017convergence,stich2018local,yu2019parallel,wang2018cooperative} and its variant \emph{federated averaging} has been shown to work well even when worker nodes have non-IID data partitions \cite{McMahan2017federated}. However, the significant communication reduction of Local SGD comes with a cost. As observed in experiments \cite{Wang2018Adaptive}, a larger number of local updates $\cp$ requires less communication but typically leads to a higher error at convergence. There is an interesting trade-off between the error-convergence and communication efficiency.

\begin{figure}[t]
    \centering
    \includegraphics[width=.3\textwidth]{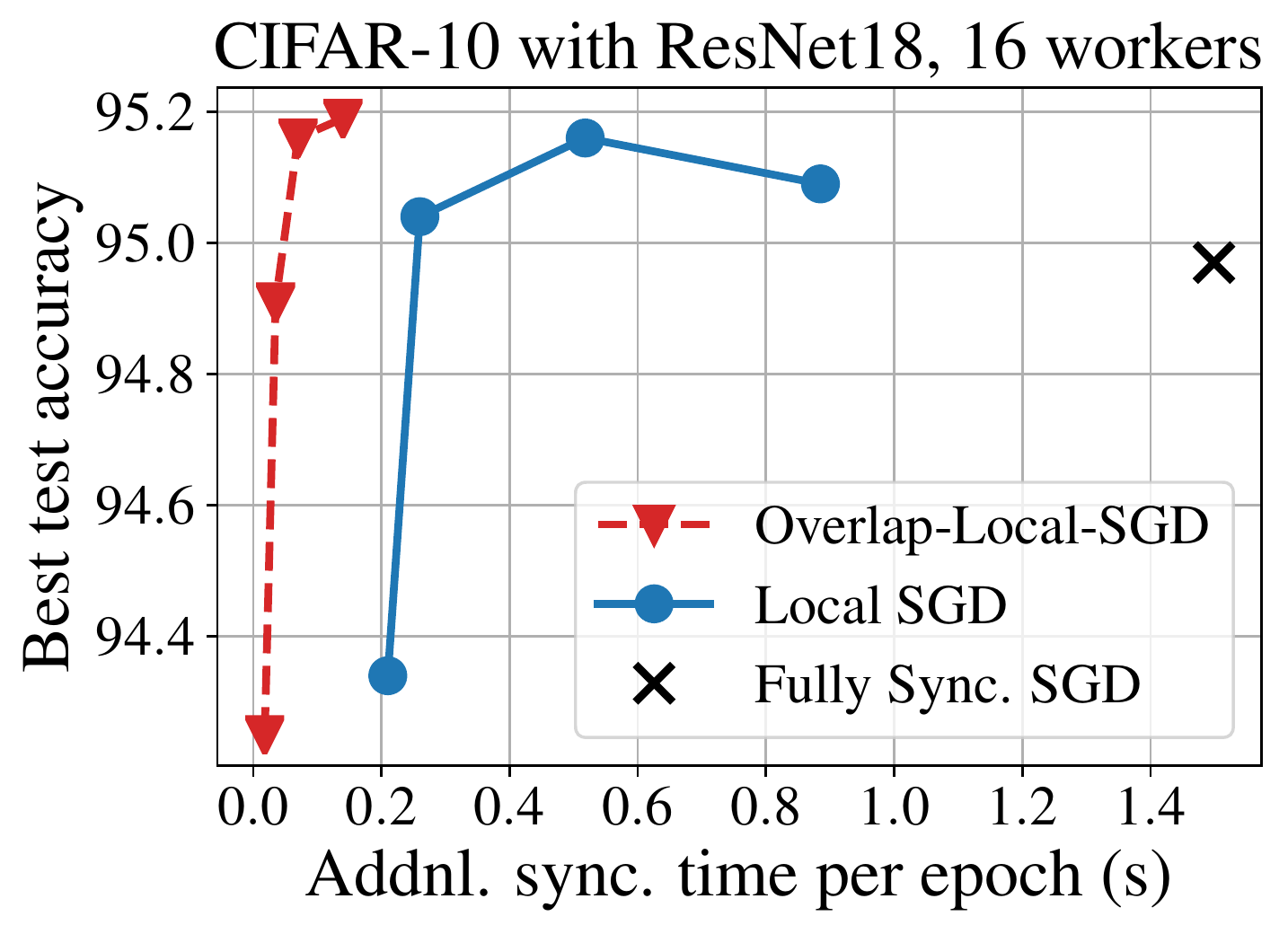}
    \caption{Error-runtime trade-off. The proposed algorithm {\alg} significantly improves the Pareto efficiency of Local SGD. Each point in the plot corresponds to a specific value of $\cp$. Note that the computation time per epoch is about $4.6$ seconds.}
    \label{fig:trade-off}
\end{figure}

In this paper, we propose a novel algorithm named \emph{\alg} that further improves the communication efficiency of Local SGD and achieves a better balance in the error-runtime trade-off. The key idea in {\alg} is introducing an anchor model on each node. {After each round of local updates}, the anchor model use another thread/process to synchronize. Thus, the communication and computation are decoupled and happen in parallel. The locally trained models achieve consensus via averaging with the synchronized anchor model instead of communicating with others. The benefits of using {\alg} is shown in \Cref{fig:trade-off}. One can observe that the additional synchronization latency per epoch is nearly negligible compared to fully synchronous SGD. By setting a proper number of updates ($\cp=1,2$), {\alg} can even achieve a higher accuracy. Extensive experiments in \Cref{sec:exps} further validate the effectiveness of {\alg} under both IID and non-IID data settings. We provide a convergence analysis in \Cref{sec:convergence} and show that the proposed algorithm can converge to a stationary point of non-convex objectives and achieve the same rate as fully synchronous SGD.

% In order to achieve a win-win in this error-runtime trade-off, previous works proposed several momentum schemes to accelerate the convergence rate with respect to iterations without sacrificing the throughput; see \cite{wang2019slowmo,yu2019linear}. In this paper, we will focus on an orthogonal direction, that is, to further reduce the communication time of Local SGD while maintaining the same error-convergence rate. 
% \begin{itemize}
%     \item We propose a new variant of Local SGD that can overlap the communication and computation so as to significantly speedup the training.
%     \item experimental results.
%     \item theoretical analysis.
% \end{itemize}

\section{Proposed Algorithm}\label{sec:algo}
\textbf{Preliminaries.} Consider a network of $\nworker$ worker nodes, each of which only has access to its local data distribution $\mathcal{D}_i$, for all $i\in\{1,\dots,\nworker\}$. Our goal is to use these $\nworker$ nodes to jointly minimize an objective function $\obj(\bx)$, defined as follows:
\begin{align}
    \obj(\bx) := \frac{1}{\nworker}\sum_{i=1}^\nworker \Exs_{\bm{s} \sim \mathcal{D}_i}\brackets{\ell(\bx;\bm{s})}
\end{align}
where $\ell(\bx;\bm{s})$ denotes the loss function for data sample $\bm{s}$, and $\bx$ denotes the parameters in the learning model. In Local SGD, each node performs mini-batch SGD updates in parallel and periodically synchronize model parameters. For the model at $i$-th worker $\bx^{(i)}$, we have
\begin{align}
    \bx_{k+1}^{(i)}=
    \begin{cases}
    \frac{1}{m}\sum_{j=1}^m [\bx_{k}^{(j)}-\lr\sg_j(\bx_k^{(j)};\xi_k^{(j)})] & (k+1) \ \text{mod} \ \cp = 0 \\
    \bx_{k}^{(i)}-\lr\sg_i(\bx_k^{(i)};\xi_k^{(i)}) & \text{otherwise}
    \end{cases}
\end{align}
where $\sg_i(\bx_k^{(i)};\xi_k^{(i)})$ represents the stochastic gradient evaluated on a random sampled mini-batch $\xi_k^{(i)}\sim\mathcal{D}_i$, and $\lr$ is the learning rate.

\textbf{\alg.} In {\alg}, each node maintains two set of model parameters: the locally trained model $\bx^{(i)}$ and an additional anchor model $\bz$, which can be considered as a stale version of the averaged local model. We omit the node index of $\bz$ since it is always synchronized and the same across all nodes. 

\begin{figure}[t]
    \centering
    \includegraphics[width=.45\textwidth]{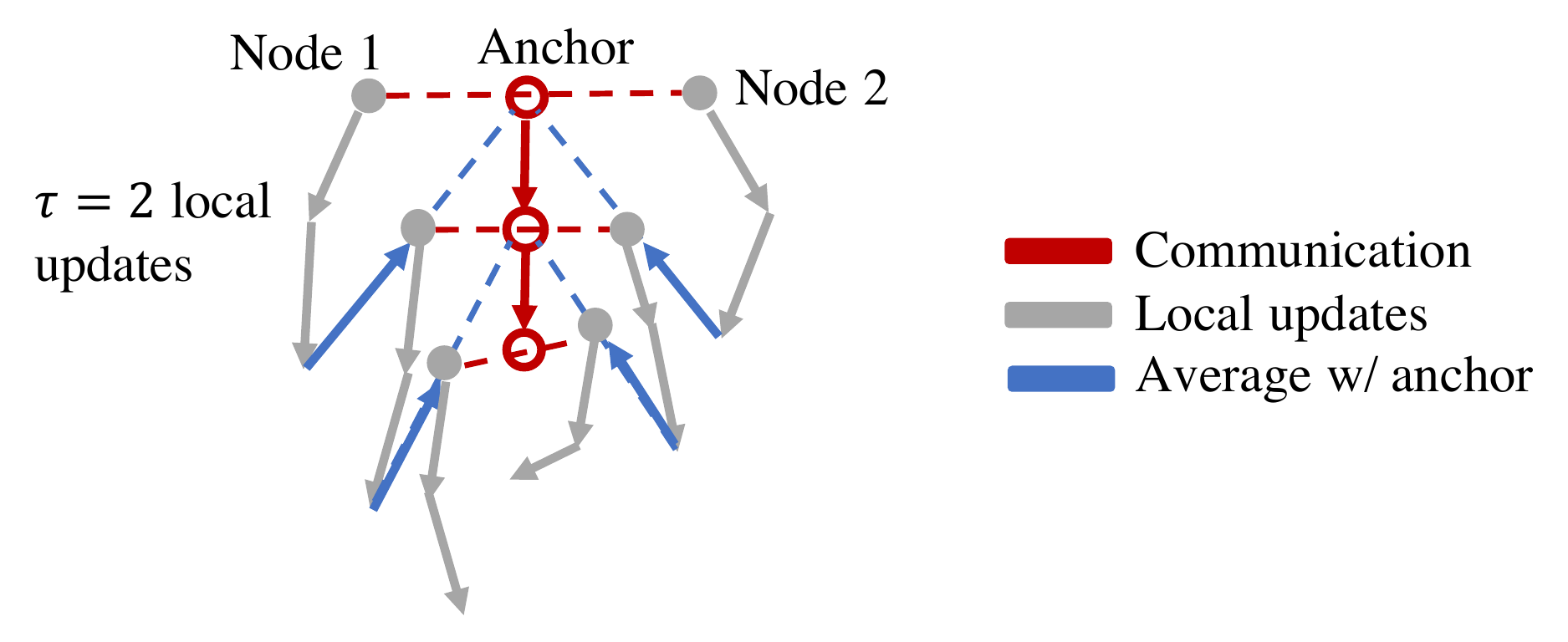}
    \caption{Example on model trajectories in the model parameter space. It is worth noting that the update of anchor is performed in parallel to the local updates of worker nodes.}
    \label{fig:model_updates}
\end{figure}

\begin{figure}[t]
    \centering
    \includegraphics[width=.45\textwidth]{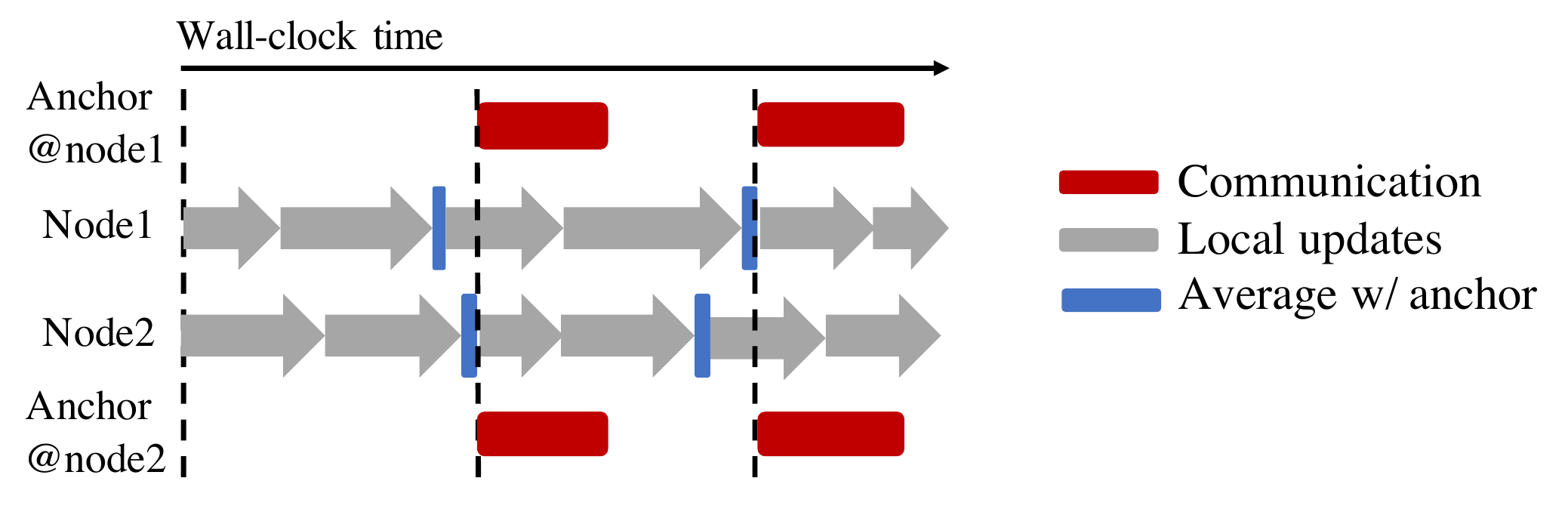}
    \caption{The corresponding execution pipeline of the example in \Cref{fig:model_updates}. There is an extra communication thread on each worker node to perform communication and update anchor models. {When the communication is done before the fastest worker completes local updates, there is no idle time in waiting for slow ones.}}
    \label{fig:timeline}
    % \vspace{-8pt}
\end{figure}

In \Cref{fig:model_updates,fig:timeline}, we present a brief illustration of {\alg}. Specifically, after every $\cp$ local updates, the updated local model $\bx^{(i)}$ will be pulled towards the anchor model. Formally, we have the following update rule for local models:
\begin{align}
    \bx^{(i)}_{k+\frac{1}{2}} & = \bx_{k}^{(i)}-\lr\sg_i(\bx_k^{(i)};\xi_k^{(i)}), \label{eqn:local_update1}\\
    \bx^{(i)}_{k+1} &= 
    \begin{cases}
    \bx^{(i)}_{k+\frac{1}{2}} - \glr(\bx^{(i)}_{k+\frac{1}{2}}-\bz_k) & (k+1) \ \text{mod} \ \cp = 0 \\
    \bx^{(i)}_{k+\frac{1}{2}} & \text{otherwise}
    \end{cases}\label{eqn:local_update2}
\end{align}
where $\glr$ is a tunable parameter. {A larger value of $\alpha$ means that the locally trained model $\bx^{(i)}$ is pulled closer to the anchor model $\bz$. Later in \Cref{sec:exps}, we will provide a empirical guideline on how to set $\alpha$ in practice. Besides, it is worth noting that the updates \Cref{eqn:local_update1,eqn:local_update2} do not involve any communication, because each node has one local copy of the anchor model.} Right after pulling back, nodes will start next round of local updates immediately. Meanwhile, another thread (or process) on each node will synchronize the current local models in parallel and store the average value into the anchor model as follows:
\begin{align}
    \bz_{k+1} &= 
    \begin{cases}
    \frac{1}{\nworker}\sum_{i=1}^\nworker\bx^{(i)}_{k+1} & (k+1) \ \text{mod} \ \cp = 0 \\
    \bz_{k} & \text{otherwise}
    \end{cases}\label{eqn:update_z}
\end{align}
{From the update rules \Cref{eqn:local_update1,eqn:local_update2,eqn:update_z}, one can observe that the anchor model $\bz_{a\cp}, a=1,2,3,\dots$ will only be used when updating $\bx^{(i)}_{(a+1)\cp}$.} As long as the parallel communication time is smaller than $\cp$ steps computation time, one can completely hide the communication latency. This can be achieved via setting a larger number of local updates $\cp$.


\textbf{Mitigating the Effect of Stragglers.} The overlap technique not only hides the communication latency but also mitigates the straggler effect. This is because the communication operations are non-blocking. {When the anchor model is updated (\ie, communication is finished) before the fastest worker completes its local updates (as shown in \Cref{fig:timeline}), all worker nodes will run independently and there is no idle time in waiting for the slow ones.}

% In order to analyze the delay model, we denote by $Y_{i,k}\sim F_Y$ the time taken by the $i$-th worker node to finish the $k$-th local step. For simplicity, $Y_{i,k}$ are assumed to be IID across nodes and iterations. In the ideal case, the $c$-th anchor update (\ie, $c$-th communication step) happens at
% \begin{align}
%     T_c = \max_{i \in \{1,\dots,m\}}\brackets{\sum_{k=0}^{c\cp-1}Y_{i,k}}.
% \end{align}
% Similarly, one can get the expression of $T_{c-1}$. Note that when the communication time $T$ is smaller than $T_c-T_{c-1}$, then there is no waiting time. This yields a threshold on the number of local updates:
% \begin{align}
%     T
%     & \leq  \Exs\brackets{\max_{i \in \{1,\dots,m\}}\brackets{\sum_{k=0}^{c\cp-1}Y_{i,k}} - \max_{i \in \{1,\dots,m\}}\brackets{\sum_{k=0}^{(c-1)\cp-1}Y_{i,k}}}. \label{eqn:cp_constraint}
% \end{align}
% The exact expression of \eqref{eqn:cp_constraint} depends on the distribution $F_Y$. In the simplest case where $Y_{i,k}$ is a constant, we have $\cp\geq T/Y$.

\textbf{Matrix-Form Update Rule.}
{In order to facilitate the theoretical analysis, here we provide an equivalent matrix-form update rule.} We define matrices $\bX_k,\bG_k \in \mathbb{R}^{d \times (\nworker+1)}$ to stack all local copies of model parameters and stochastic gradients:
\begin{align}
    \bX_k &= [\bx_k^{(1)},\dots,\bx_k^{(\nworker)},\bz_k], \\
    \bG_k &= [\sg_1(\bx_k^{(1)};\xi_k^{(1)}), \dots, \sg_\nworker(\bx_k^{(\nworker)};\xi_k^{(\nworker)}),\bm{0}].
\end{align}
Then, the update rule of {\alg} can be written as
\begin{align}
    \bX_{k+1} &= [\bX_k - \lr \bG_k]\bW_k,\label{eqn:update}
\end{align}
where $\bW_k \in \mathbb{R}^{(\nworker+1)\times (\nworker+1)}$ represents the mixing pattern between local models and the anchor model, which is defined as follows:
\begin{align}
    \bW_k &= 
    \begin{cases}
    \begin{bmatrix}
    (1-\glr)\bm{I} & (1-\glr)\one_\nworker/m \\
    \glr\one_\nworker\tp & \glr
    \end{bmatrix} & (k+1) \ \text{mod} \ \cp = 0 \\
    \bm{I} & \text{Otherwise}.
    \end{cases} \label{eqn:W}
\end{align}
Note that $\bW_k$ is a column-stochastic matrix, {unlike previous analyses in distributed optimization literature \cite{Nedic2018network,wang2018cooperative,assran2018stochastic}, which require $\bW_k$ to be doubly- or row-stochastic.} 

%This is also the key difference between overlap local SGD and elastic averaging SGD (EASGD) \cite{Zhang2015elasticsgd}.

\textbf{Momentum Variant.}
Momentum has been widely used to improve the optimization and generalization performance of SGD, especially when training deep neural networks \cite{sutskever2013importance}. Inspired by the distributed momentum scheme proposed in \cite{wang2019slowmo}, {\alg} adopts a two-layer momentum structure. To be specific, the local updates on each node use common Nesterov momentum and the momentum buffer is updated only using the local gradients. Moreover, the anchor model also updates in a momentum style. When $(k+1) \ \text{mod}\ \cp =0$, we have
\begin{align}
    \bm{v}_{k+1} &= \beta \bm{v}_k +\left(\frac{1}{m}\sum_{i=1}^m \bx_{k+1}^{(i)} - \bz_k \right), \\
    \bz_{k+1} &= \bz_{k} + \bm{v}_{k+1}
\end{align}
where $\bm{v}_k$ is the momentum buffer for anchor model and $\beta$ denotes the momentum factor. {When $\beta=0$, the algorithm reduces to the vanilla version as \Cref{eqn:update_z}}.

%\GJ{'Eliminate' is too strong of a word. We don't completely eliminate but just mitigate the effect of stragglers.}

% \begin{algraging SGD (EASGD) \cite{zha}ori Howeverthm}
%     \DontPrintSemicolon
%     \SetKwInput{Input}{Input}
%     \SetAlgoLined
%     \LinesNumbered
%     \Input{Initial local model $\bz_0=\bx_0^{(i)}, \forall i \in\{1,2,\dots,m\}$; Objective function $\obj_i(\bx)$; Learning rate $\lr$; Number of Local updates $\cp$.}
%      \For{$k \in \{0,1,\dots,K-1\}$}{
%         Local update: $\bx_{k+\frac{1}{2}}^{(i)} = \bx_{k}^{(i)} - \lr\sg_i(\bx_k^{(i)};\xi_k^{(i)})$ \;
%         \eIf{$(k+1) \mod \cp = 0$}{
%         Move towards anchor: $\bx_{k+1}^{(i)} = \bx_{k+\frac{1}{2}}^{(i)} - \glr(\bx_{k+\frac{1}{2}}^{(i)}-\bz)$\;
%         }{
%         $\bx_{k+1}^{(i)} = \bx_{k+\frac{1}{2}}^{(i)}$\;
%         }
%      Update slow model: $\bz_t = \bz_{t-1} + \glr(\bx_{t,\cp}-\bx_{t,0})$\;
%      }
%      {\bf Return} slow model $\bz_{T-1}$
%      \caption{Overlap Local SGD}
%      \label{algo:anchor-agent}
% \end{algorithm}

\section{Related Works}\label{sec:related}
The idea of pulling back locally trained models towards an anchor model is inspired by elastic averaging SGD (EASGD) \cite{Zhang2015elasticsgd}, which allows some slack between local models by adding a proximal term to the objective function. The convergence guarantee of EASGD under non-convex objectives has not been established until our recent work \cite{wang2018cooperative}. However, in EASGD, the anchor and local models are updated in a symmetric manner (\ie, mixing matrix $\bW_k$ in \eqref{eqn:update} should be symmetric and doubly-stochastic). EASGD naturally allows overlap of communication and computation, but the original paper \cite{Zhang2015elasticsgd} did not observe and utilize this advantage to reduce communication delays. %\GJ{Avoid saying 'Did not realize'. Can instead say -- EASGD naturally allows overlap of communication and computation, but the original paper \cite{Zhang2015elasticsgd} did not observe and utilize this advantage to reduce communication delays}

There also exist other techniques that can decouple communication and computation in Local SGD. In \cite{Shen2019FasterDD}, the authors propose to apply the local updates to an averaged model which is $\cp$-iterations before. Their proposed algorithm CoCoD-SGD can achieve the same runtime benefits as {\alg}. Nonetheless, later in \Cref{sec:exps}, we will show that, {\alg} consistently reaches comparable or even higher test accuracy than CoCoD-SGD given the same $\cp$. In a concurrent work \cite{wang2019osp}, the authors develop a similar method to CoCoD-SGD.

\section{Experimental Results}\label{sec:exps}

\textbf{Experimental setting.} The experimental analysis is performed on CIFAR-10 image classification task \cite{krizhevsky2009cifar}. We train a ResNet-18 \cite{he2016deep} for $300$ epochs following the exactly same training schedule as \cite{vogels2019powersgd}. That is, the mini-batch size on each node is $128$ and the base learning rate is $0.1$, decayed by $10$ after epoch $150$ and $250$. The first 5 epoch uses the learning rate warmup schedule as described in \cite{goyal2017accurate}. There are total $16$ computing nodes connected via $40$ Gbps Ethernet, each of which is equipped with one NVIDIA Titan X GPU. The training data is evenly partitioned across all nodes and \emph{not shuffled} during training. The algorithms are implemented in PyTorch \cite{paszkepytorch} and NCCL communication backend. The code is available at: \url{https://github.com/JYWa/Overlap_Local_SGD}.

In {\alg}, the momentum factor of the anchor model is set to $\beta=0.7$, following the convention in \cite{wang2019slowmo}. For different number of local updates $\cp$, we tune the value of pullback parameter $\glr$. It turns out that in the considered training task, for $\cp\geq2$, $\glr=0.6$ consistently yields the best test accuracy at convergence. In intuition, a larger value of $\glr$ may enable a larger base learning rate. We believe that if one further tune the base learning rate and the momentum factor, the performance of {\alg} will be further improved. For example, in our setting, when $\cp=1$, then $\glr=0.5$ and base learning rate $0.15$ gives the highest accuracy.

\begin{figure*}[!hbtp]
    \centering
    \begin{subfigure}{.3\textwidth}
    \centering
    \includegraphics[width=\textwidth]{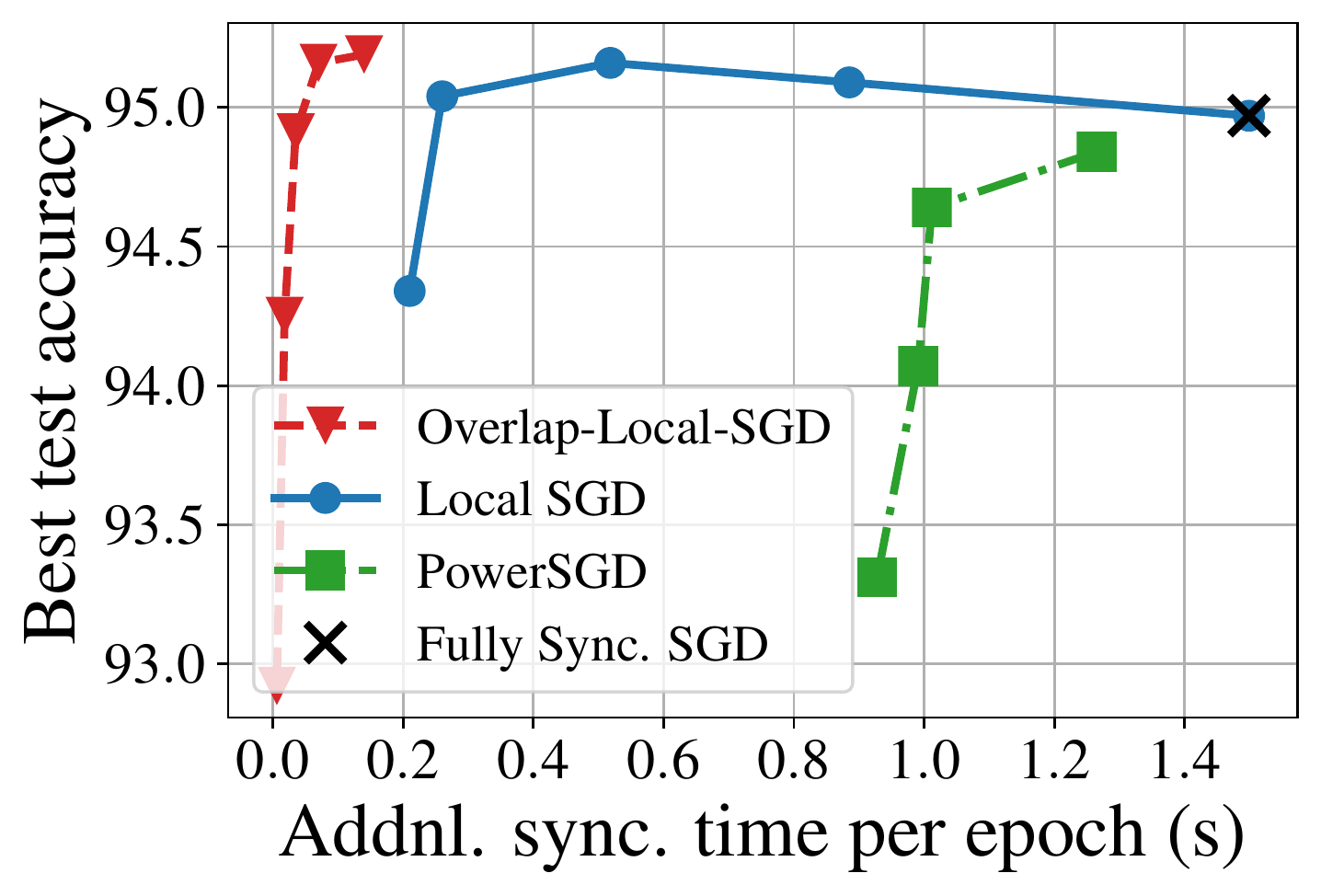}
    \caption{Error-runtime trade-off.}
    \end{subfigure}%
    ~
    \begin{subfigure}{.3\textwidth}
    \centering
    \includegraphics[width=\textwidth]{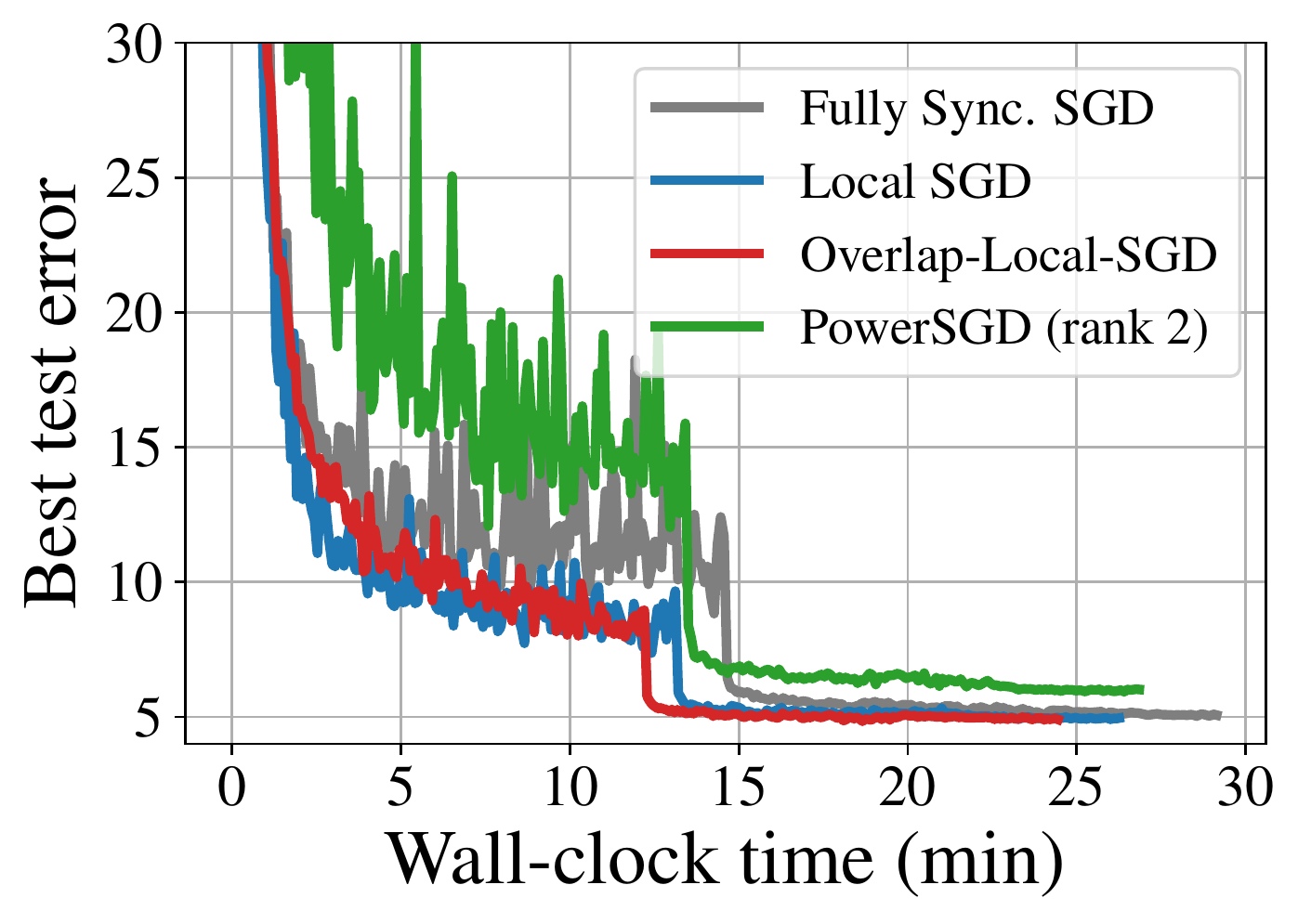}
    \caption{Validation curves.}
    \end{subfigure}%
    ~
    \begin{subfigure}{.3\textwidth}
    \centering
    \includegraphics[width=\textwidth]{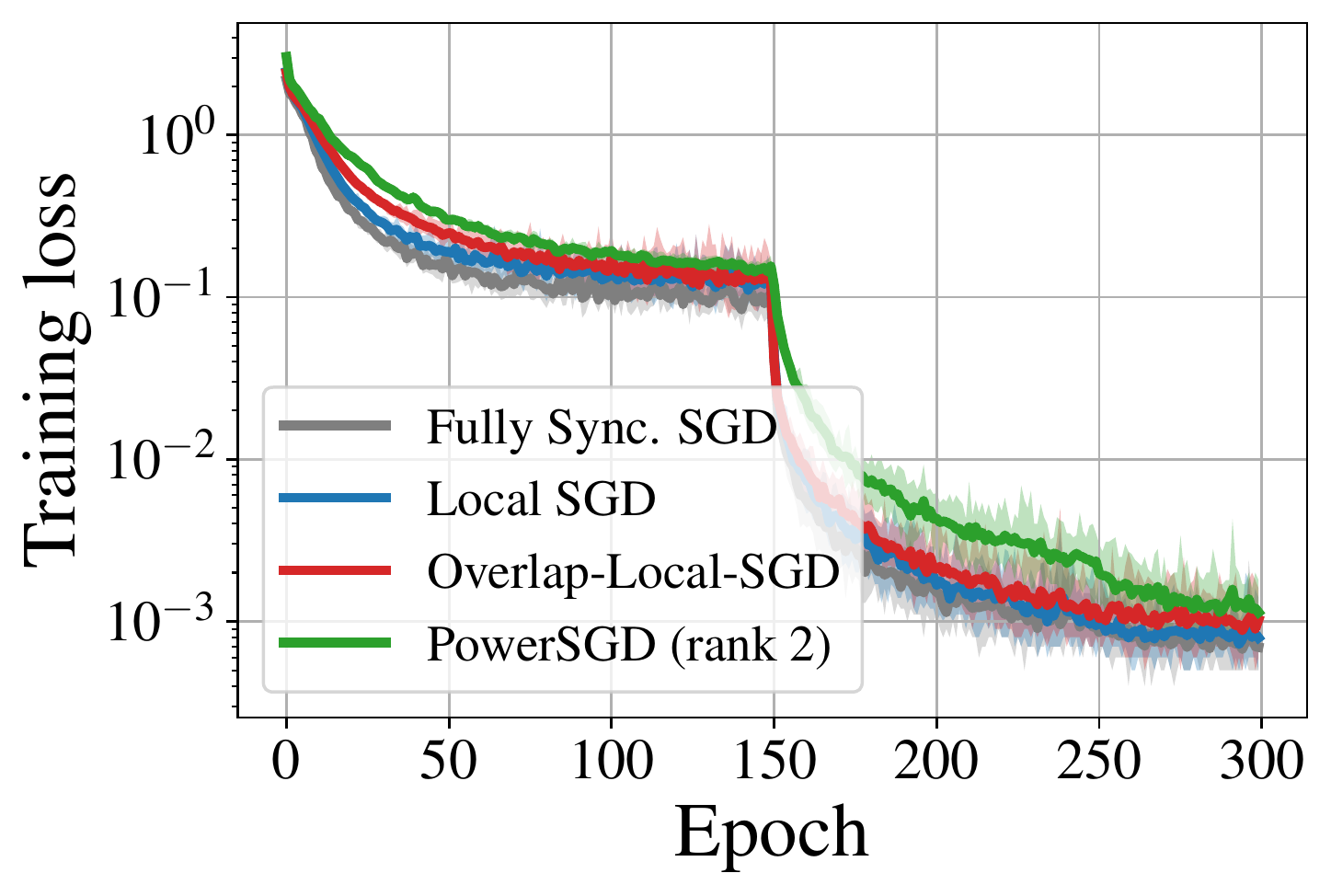}
    \caption{Training curves.}
    \end{subfigure}
    \caption{Comparison of communication-efficient SGD methods in \textbf{IID} data partition setting. In (a), the number of local updates of Local SGD method is taken from $\{1,2,4,8,24\}$. In (b) and (c), we fix $\cp=2$.}
    \label{fig:IID}
\end{figure*}

\begin{figure*}[t]
    \centering
    \begin{subfigure}{.3\textwidth}
    \centering
    \includegraphics[width=\textwidth]{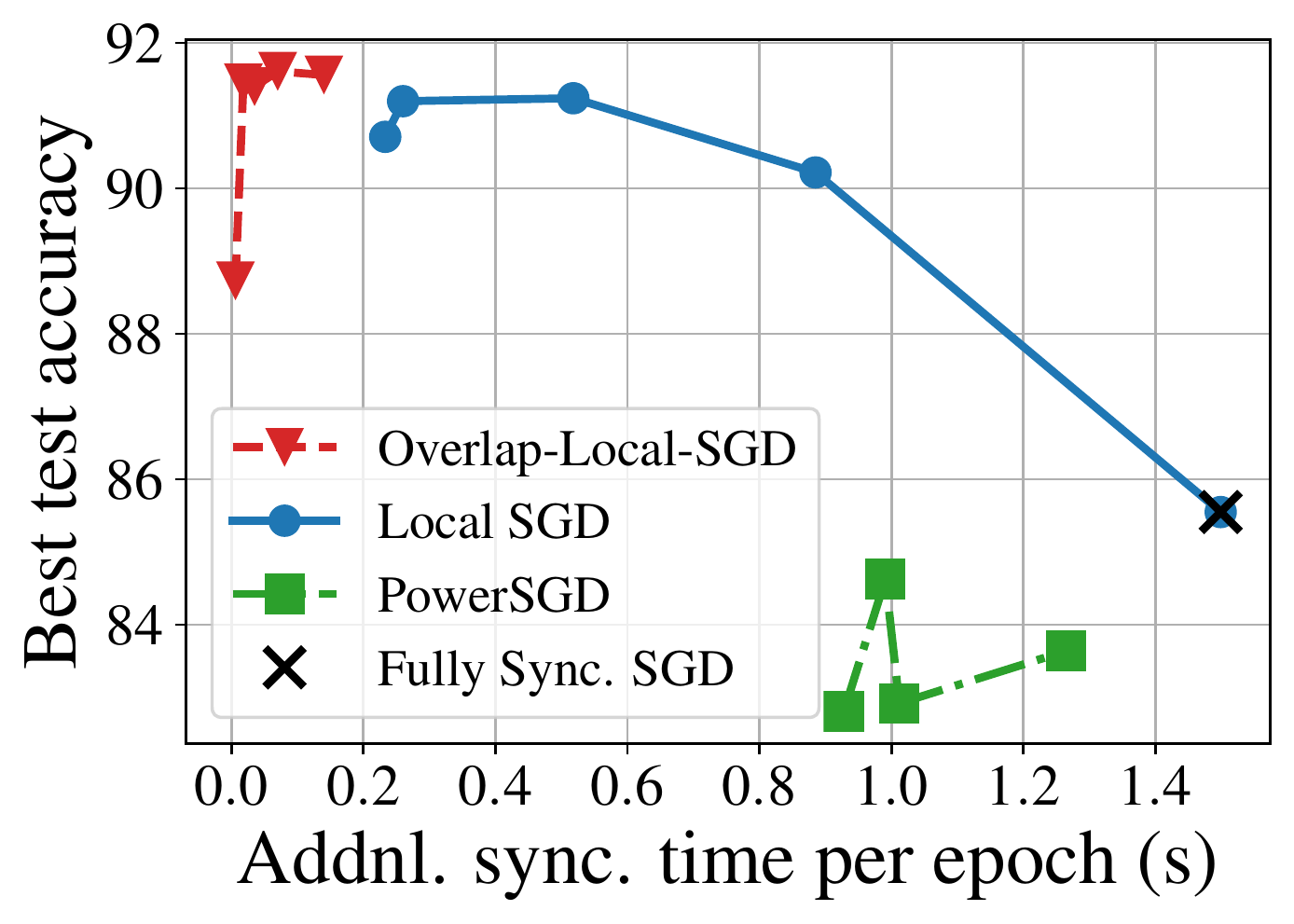}
    \caption{Error-runtime trade-off.}
    \end{subfigure}%
    ~
    \begin{subfigure}{.3\textwidth}
    \centering
    \includegraphics[width=\textwidth]{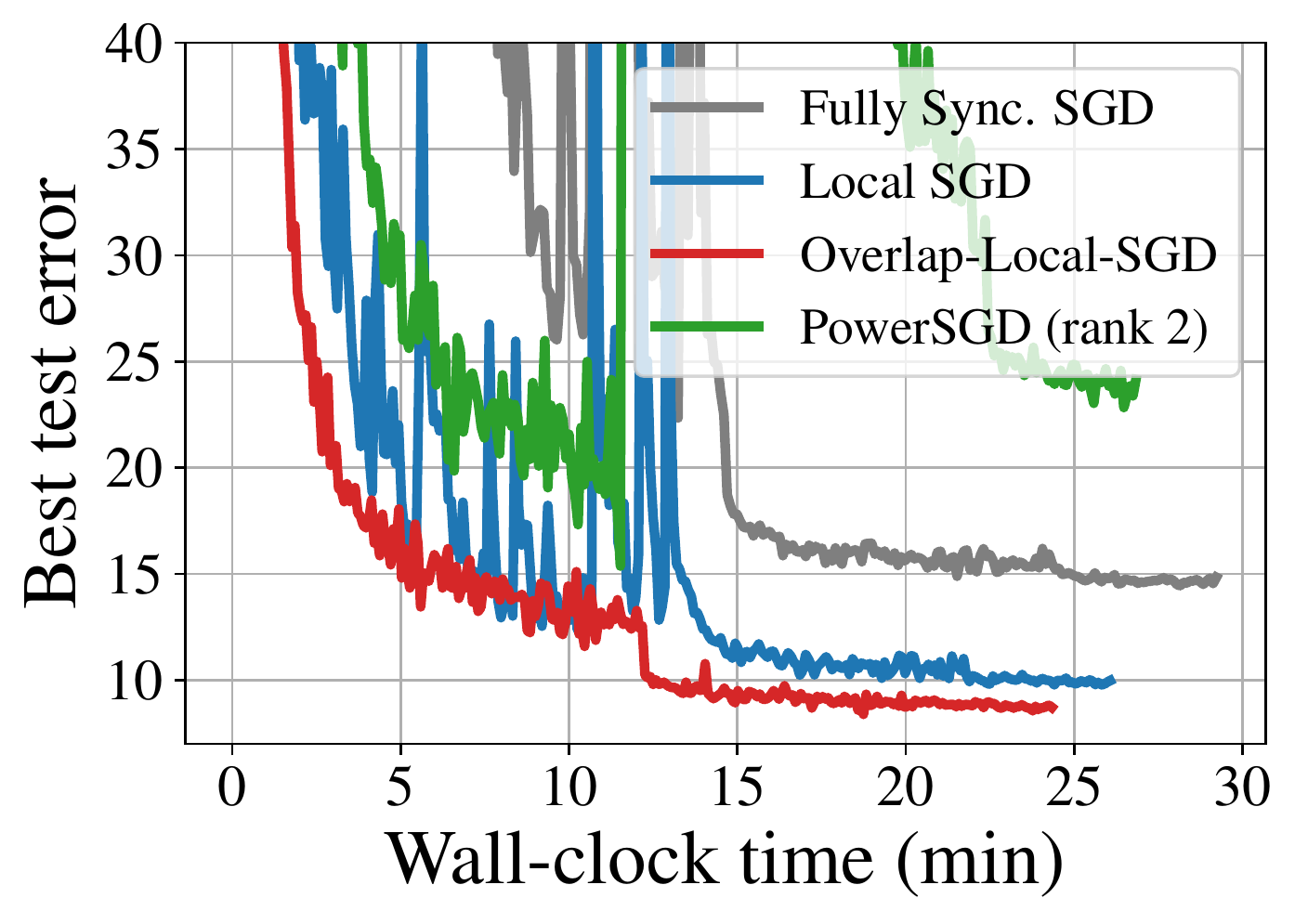}
    \caption{Validation curves.}
    \end{subfigure}%
    ~
    \begin{subfigure}{.3\textwidth}
    \centering
    \includegraphics[width=\textwidth]{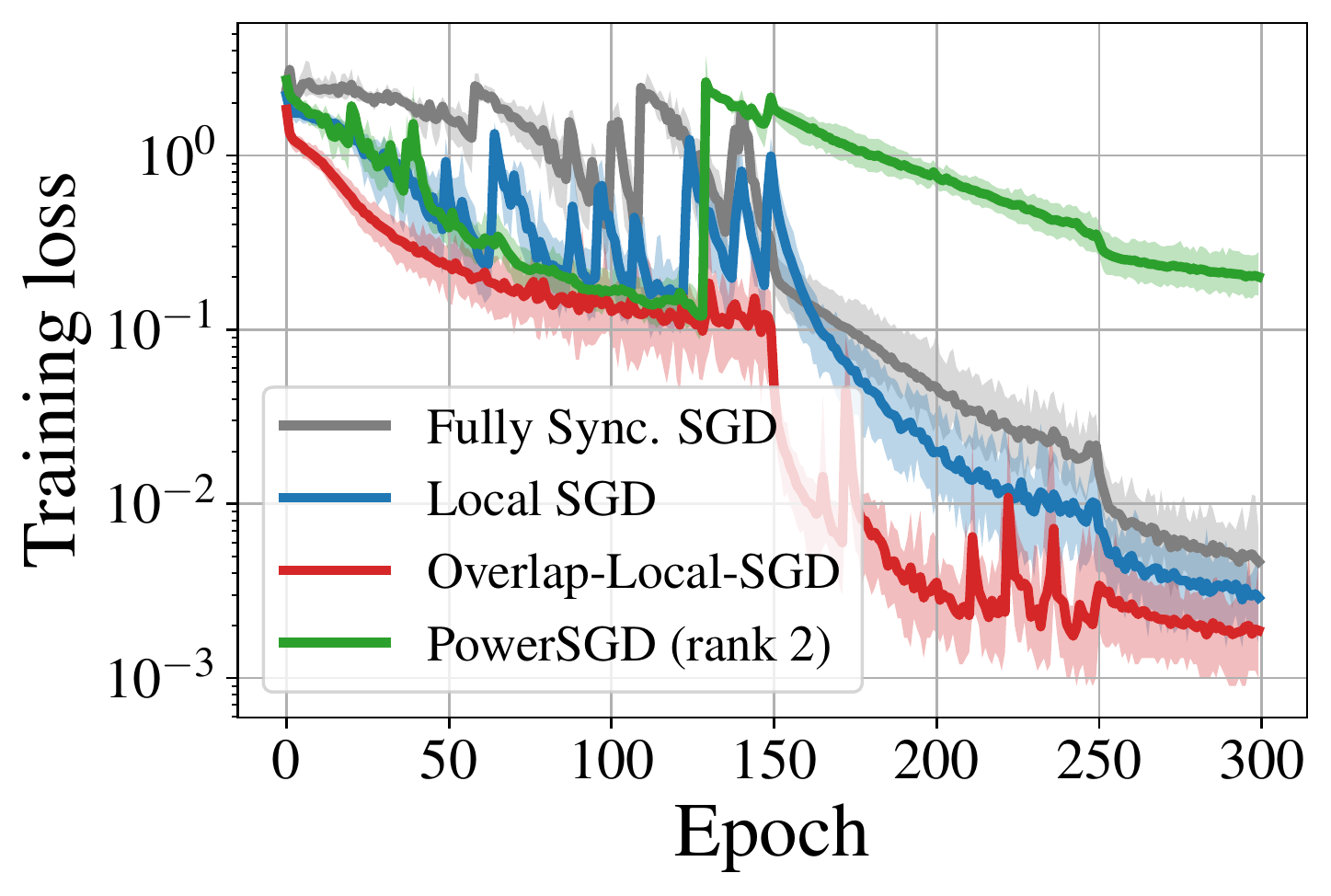}
    \caption{Training curves.}
    \end{subfigure}
    \caption{Comparison of communication-efficient SGD methods in \textbf{non-IID} data partition setting. In (a), the number of local updates of Local SGD method is taken from $\{1,2,4,8,24\}$. In (b) and (c), we fix $\cp=2$. {\alg} is much more stable than other methods.}
    \label{fig:NIID}
\end{figure*}

\textbf{Negligible Communication Cost.}
We first examine the effectiveness of the overlap technique. As shown in \Cref{fig:IID}(a), {\alg} significantly outperforms all other methods. Given a target final accuracy, {\alg} incurs nearly negligible additional latency compared to fully synchronous SGD ($0.1$s versus $1.5$s per epoch). When $\cp=2$, {\alg} reduces the communication-to-computation ratio from $34.6\%$ to $1.5\%$, while maintaining roughly the same loss-versus-iterations convergence as fully synchronous SGD (see \Cref{fig:IID}(c)). The superiority of {\alg} will be further magnified when using a slow inter-connection (\eg, $10$ Gbps) or a larger neural network (\eg, transformer \cite{ott2018scaling}).

Compressing or quantizing the exchanged gradients among worker nodes is another communication-efficient training method, which is extensively studied in recent literature. Here, we choose PowerSGD \cite{vogels2019powersgd}, which is the state-of-the-art gradient compression algorithm, as another baseline to compare with. In \Cref{fig:IID}, the rank of PowerSGD ranges from $\{1,2,4,8\}$ (lower means higher compression ratio). When the rank is 1 (the lowest), PowerSGD can compress the transferred gradient by $243\times$. However, even in this extreme case, the additional synchronization latency of PowerSGD is still much higher than Local SGD methods. The reason is that the nodes cost some time to establish the handshakes. Compression techniques cannot reduce this part of communication overhead, and also introduce non-negligible encoding and decoding latency.

\textbf{Higher Accuracy than Other Local SGD Variants.} As discussed in \Cref{sec:related}, EASGD (and its momentum version EAMSGD \cite{Zhang2015elasticsgd}) also involve(s) a similar `pullback' mechanism as {\alg}. And CoCoD-SGD proposed in \cite{Shen2019FasterDD} can decouple communication and computation as well. In \Cref{tab:comparison_iid}, we empirically compare the performance of these Local SGD variants. The results show that given a fixed number of $\cp$, {\alg} always achieves the best test accuracy among all methods, and EAMSGD has significant worse performance than others.  

\begin{table}[t]
    \centering
    \begin{tabular}{c c c c c c}\toprule
    Algorithm & $\cp=1$ & $\cp=2$ & $\cp=8$ & $\cp=24$\\\midrule
    CoCoD-SGD &  $94.98\%$ & $94.99\%$ & $94.05\%$ & $92.54\%$\\
    EAMSGD    &  $94.51\%$ & $93.89\%$ & $92.43\%$ & $89.93\%$ \\
    Ours & $\bm{95.19\%}$ & $\bm{95.16\%}$ & $\bm{94.25\%}$ & $\bm{92.92\%}$ \\\bottomrule 
    \end{tabular}
    \caption{Comparison of Local SGD variants in \textbf{IID} data partition setting. As a reference, fully synchronous SGD achieves a test accuracy of $94.97\%$. The corresponding training loss curves can be found in Appendix B.}
    \label{tab:comparison_iid}
\end{table}

\begin{table}[t]
    \centering
    \begin{tabular}{c c c c c c}\toprule
    Algorithm & $\cp=1$ & $\cp=2$ & $\cp=8$ & $\cp=24$\\\midrule
    CoCoD-SGD &  $91.50\%$ & $\bm{91.67\%}$ & Diverges & Diverges \\
    EAMSGD    &  $91.38\%$ & $91.12\%$ & $88.88\%$ & $85.59\%$ \\
    Ours & $\bm{91.56\%}$ & $91.61\%$ & $\bm{91.45\%}$ & $\bm{88.73\%}$ \\\bottomrule 
    \end{tabular}
    \caption{Comparison of Local SGD variants in \textbf{Non-IID} data partition setting. As a reference, fully synchronous SGD achieves a test accuracy of $85.88\%$. The hyper-parameter choices are identical to the IID case.}
    \label{tab:comparison_niid}
\end{table}

\textbf{Non-IID Data Partitions Setting.} We further validate the effectiveness of {\alg} in a non-IID data partitions setting. In particular, each node is assigned with $3125$ training samples, $2000$ of which are belong to one class. Thus, the training data on each node is highly skewed. In \Cref{fig:NIID}, observe that both fully synchronous SGD and Local SGD are pretty unstable in this case. {\alg} not only reduces the total training time but also yields better convergence in terms of error-versus-iterations (see \Cref{fig:NIID}(c)). Compared to CoCoD-SGD (see \Cref{tab:comparison_niid}), {\alg} still can achieve comparable test accuracy and overcome the divergence issue when $\cp$ is large.

% \begin{figure}[b]
%     \centering
%     \includegraphics[width=.3\textwidth]{overlap_results.pdf}
%     \caption{Example of the error-runtime trade-off. The proposed algorithm {\alg} significantly improves the Pareto efficiency of communication-efficient SGD. Note that the computation time per epoch is about $0.0$s [TBA]. PowerSGD \cite{vogels2019powersgd} is the state-of-the-art gradient compression technique.}
%     \label{fig:trade-off}
% \end{figure}

\section{Convergence Analysis}\label{sec:convergence}
In this section, we will provide a convergence guarantee for {\alg} under non-convex objectives, which are common for deep neural networks. The analysis is based on the following assumptions:
\begin{enumerate}
    \item Each local objective function $\obj_i(\bx):=\Exs_{\bm{s} \sim \mathcal{D}_i}\brackets{\ell(\bx;\bm{s})}$ is L-smooth: $\vecnorm{\tg_i(\bx)-\tg_i(\by)}\leq \lip \vecnorm{\bx - \by}, \forall i\in [1,\nworker]$.
    \item The stochastic gradients are unbiased estimators of local objectives' gradients, \ie, $\Exs_{\xi\sim \mathcal{D}_i}[\sg_i(\bx;\xi)]=\tg_i(\bx)$.
    \item The variance of stochastic gradients is bounded by a non-negative constant: $\Exs_{\xi\sim \mathcal{D}_i}[\vecnorm{\sg_i(\bx;\xi)-\tg_i(\bx)}^2]\leq \vbnd$.
    \item The average deviation of local gradients is bounded by a non-negative constant: $\frac{1}{\nworker}\sum_{i=1}^\nworker \vecnorm{\tg_i(\bx)-\tg(\bx)}^2\leq \kappa^2$.
\end{enumerate}

Formally, we have the following theorem. It can guarantee that {\alg} converges to stationary points of non-convex objective functions.
\begin{thm}\label{thm:main}
Suppose all local models and anchor model are initialized at the same point $\bx_0^{(i)}=\bz_0$ for all $i \in \{1,\dots,\nworker\}$. Under Assumptions 1 to 4, if the learning rate is set as $\lr=\frac{1}{\lip}\sqrt{\frac{\nworker}{K}}$, and the total iterations $K$ satisfies $K \geq 60\nworker\cp^2/\glr^2$, then we have
\begin{align}
    \frac{1}{K}\sum_{k=0}^{K-1}\Exs\brackets{\vecnorm{\tg(\by_k)}^2} 
    \leq& \frac{4\lip[\obj(\by_0)-\obj_\text{inf}]}{(1-\glr)\sqrt{\nworker K}} + \frac{2(1-\glr)\vbnd}{\sqrt{\nworker K}} + \nonumber \\
        & \frac{2\nworker\vbnd}{K}\brackets{\frac{2}{(2-\glr)\glr}\cp-1} + \frac{2\nworker\cp^2\kappa^2}{\glr^2 K} \\
    =& \mathcal{O}\parenth{\frac{1}{\sqrt{\nworker K}}} + \mathcal{O}\parenth{\frac{1}{K}}.
\end{align}
where $\by_k = (1-\glr)\sum_{i=1}^\nworker \bx_k^{(i)} + \glr \bz_k$ and $\obj_\text{inf}$ is the lower bound of the objective value.
\end{thm}
Due to space limitation, please refer to Appendix A for the proof details. Briefly, the proof technique is inspired by \cite{wang2018cooperative}. The key challenge is that the mixing matrix of {\alg} is column-stochastic instead of doubly- or row-stochastic \cite{Nedic2018network}. It is worth highlighting that the analysis can be generalized to other column stochastic matrices rather than the specific form given in \Cref{eqn:W}. \Cref{thm:main} also shows that when the learning rate is configured properly and the total iterations $K$ is sufficiently large, the error bound of {\alg} will be dominated by $1/\sqrt{\nworker K}$, matching the same rate as fully synchronous SGD.

\section{Conclusions}
In this paper, we propose a novel distributed training algorithm named {\alg}. It allows workers to perform local updates and overlaps the local computation and communication. Experimental results on CIFAR-10 show that {\alg} can achieve the best error-runtime trade-off among multiple popular communication-efficient training methods, such as Local SGD and PowerSGD. Moreover, when worker nodes have non-IID data partitions, {\alg} not only reduces the total runtime but also converges faster than other methods. We further prove that {\alg} can converge to stationary points of smooth and non-convex objective functions. While our experiments and analysis only focus on image classification and SGD, the key idea of {\alg} can be easily extended to other training task and first-order optimization algorithms, such as Adam \cite{kingma2014adam} for neural machine translation \cite{ott2018scaling}.

\newpage
\bibliographystyle{IEEEbib}
\bibliography{decentralized_opt}

\newpage
\appendix
\section{Proof of Theorem 1}
Recall the update rule of Overlap-Local-SGD:
\begin{align}
    \bX_{k+1} &= [\bX_k - \lr \bG_k]\bW_k. \label{eqn:update2}
\end{align}
Matrix $\bW_k$ is column-stochastic as defined in \eqref{eqn:W}. To be specific,
\begin{align}
    \bW_k &= 
    \begin{cases}
    \bP & (k+1) \ \text{mod} \ \cp = 0 \\
    \bm{I} & \text{Otherwise}.
    \end{cases}
\end{align}
where matrix $\bP \in \mathbb{R}^{(\nworker+1)\times(\nworker+1)}$ is defined as
\begin{align}
    \bP = 
    \begin{bmatrix}
    (1-\glr)\bm{I} & (1-\glr)\one_\nworker/m \\
    \glr\one_\nworker & \glr
    \end{bmatrix}.
\end{align}
There must be a vector $\bv\in\mathbb{R}^{\nworker+1}$ such that $\bP\bv = \bv$ and hence $\bW_k\bv = \bv$. In particular, for the matrix given in \eqref{eqn:W}, $\bv=[(1-\glr)\one/m, \glr]$. Multiplying $\bv$ on both sides of \eqref{eqn:update2}, we have
\begin{align}
    \bX_{k+1}\bv 
    &= \bX_k\bv - \lr\bG_k\bv \\
    &= \bX_k\bv - \frac{(1-\glr)\lr}{\nworker}\sum_{i=1}^\nworker\sg_i(\bx_k^{(i)};\xi_k^{(i)}).
\end{align}
For the ease of writing, we introduce a virtual sequence $\by_k := \bX_k\bv=(1-\glr)\sum_{i=1}^\nworker\bx_k^{(i)}/\nworker+\glr\bz_k$, and define effective learning rate as $\efflr:=(1-\glr)\lr$. Consequently, we get an equivalent vector-form update rule for {\alg} as follows:
\begin{align}
    \by_{k+1} = \by_k - \efflr\frac{1}{\nworker}\sum_{i=1}^\nworker\sg_i(\bx_k^{(i)};\xi_k^{(i)}).
\end{align}
Then, we can directly apply Lemma 3 in \cite{wang2018cooperative} and obtain the following (when $\efflr\lip \leq 1$)
\begin{align}
    \frac{1}{K}\sum_{k=0}^{K-1}\Exs\brackets{\vecnorm{\tg(\by_k)}^2} 
    &\leq \frac{2[\obj(\by_0)-\obj_\text{inf}]}{\efflr K} + \frac{\efflr\lip\vbnd}{\nworker} + \nonumber \\
        & \frac{\lip^2}{K\nworker}\sum_{k=0}^{K-1}\sum_{i=1}^\nworker\Exs\brackets{\vecnorm{\by_k-\bx_k^{(i)}}^2}. \label{eqn:res0}
\end{align}
Note that
\begin{align}
    \sum_{i=1}^\nworker\vecnorm{\by_k-\bx_k^{(i)}}^2
    \leq& \sum_{i=1}^\nworker\vecnorm{\by_k-\bx_k^{(i)}}^2 + \vecnorm{\by_k-\bz_k}^2 \\
    =& \fronorm{\bX_k\parenth{\bm{I}-\bv\one\tp}}^2.
\end{align}
According to the update rule \Cref{eqn:update2} and repeatedly using the fact $\bW_k\bv = \bv, \one\tp \bW_k = \one\tp$ and $\bv\tp\one=1$, we have
\begin{align}
    &\bX_k\parenth{\bm{I}-\bv\one\tp} \nonumber \\
    =& \parenth{\bX_{k-1}-\efflr\bG_{k-1}}\bW_k\parenth{\bm{I}-\bv\one\tp}\\
    =& \bX_{k-1}\parenth{\bW_{k-1} - \bv\one\tp} - \efflr\bG_{k-1}\parenth{\bW_{k-1} - \bv\one\tp} \\
    =& \bX_0\parenth{\prod_{j=0}^{k-1}\bW_j - \bv\one\tp} - \efflr \sum_{j=0}^{k-1} \bG_j\parenth{\prod_{s=j}^{k-1}\bW_s - \bv\one\tp} \\
    =& \bx_0\one\tp\parenth{\prod_{j=0}^{k-1}\bW_j - \bv\one\tp} - \efflr \sum_{j=0}^{k-1} \bG_j\parenth{\prod_{s=j}^{k-1}\bW_s - \bv\one\tp} \\
    =& -\efflr \sum_{j=0}^{k-1} \bG_j\parenth{\prod_{s=j}^{k-1}\bW_s - \bv\one\tp}.
\end{align}
Therefore,
\begin{align}
    \sum_{i=1}^\nworker\vecnorm{\by_k-\bx_k^{(i)}}^2
    \leq& \efflr^2 \fronorm{\sum_{j=0}^{k-1} \bG_j\parenth{\prod_{s=j}^{k-1}\bW_s - \bv\one\tp}}^2. \label{eqn:deviation2}
\end{align}
Here we observe that the analysis of {\alg} is very similar to the general analysis in \cite{wang2018cooperative}. The difference is that we only require $\bW_k$ to be column-stochastic instead of symmetric and doubly-stochastic. As a result, $\prod_{s=0}^\infty \bW_s$ converges to $\bv\one\tp$ rather than $\one\one\tp/\nworker$. Then, one can directly re-use the intermediate results in \cite{wang2018cooperative} and get that
\begin{align}
    &\frac{1}{K\nworker}\sum_{k=0}^{K-1}\sum_{i=1}^\nworker\Exs\brackets{\vecnorm{\by_k-\bx_k^{(i)}}^2} \nonumber \\
    \leq& \lr^2\vbnd\parenth{\frac{1+\zeta^2}{1-\zeta^2}\cp-1} + \nonumber \\
        & \frac{\lr^2\cp^2}{1-\spgap} \parenth{\frac{2\spgap^2}{1+\spgap}+\frac{2\spgap}{1-\spgap} +  \frac{\cp-1}{\cp}}\frac{1}{K\nworker}\sum_{k=0}^{K-1}\sum_{i=1}^\nworker\Exs\brackets{\vecnorm{\tg_i(\bx_k^{(i)})}^2} \label{eqn:deviation3}
\end{align}
where $\spgap:=\opnorm{\bP-\bv\one\tp}$. In order to guarantee that the upper bound  \Cref{eqn:deviation3} makes sense, $\spgap$ should be strictly smaller than $1$. Now, we are going to provide an analytical expression of $\spgap$ for the specific $\bP$ chosen in {\alg}. One can also design other forms of $\bP$ as long as $\spgap<1$.

Observe that the matrix $\bP$ can be decomposed into two parts:
\begin{align}
    \bP = (1-\glr)\bm{A} + \glr \bm{b}\one\tp \label{eqn:p_decomp}
\end{align}
where $\bm{b} = [0,\dots,0,1]\in\mathbb{R}^{\nworker+1}$ and
\begin{align}
    \bm{A}=
    \begin{bmatrix}
    \bm{I} & \one_\nworker/m \\
    \bm{0} & 0
    \end{bmatrix}.
\end{align}
Both $\bm{A}$ and $\bm{b}\one\tp$ are column-stochastic matrix. Actually, the formulation \Cref{eqn:p_decomp} is widely used in the PageRank algorithm \cite{page1999pagerank}. It is proved in \cite{haveliwala2003second} that: $\spgap=\opnorm{\bP-\bv\one\tp}\leq (1-\glr)$. Plugging the expression of $\spgap$ into \Cref{eqn:deviation3} and further relaxing the upper bound, we obtain:
\begin{align}
    &\frac{1}{K\nworker}\sum_{k=0}^{K-1}\sum_{i=1}^\nworker\Exs\brackets{\vecnorm{\by_k-\bx_k^{(i)}}^2} \nonumber \\
    \leq& \lr^2\vbnd\brackets{\frac{2}{(2-\glr)\glr}\cp-1} +\nonumber \\
        & \frac{5\lr^2\cp^2}{\glr^2}\frac{1}{K\nworker}\sum_{k=0}^{K-1}\sum_{i=1}^\nworker\Exs\brackets{\vecnorm{\tg_i(\bx_k^{(i)})}^2}. \label{eqn:deviation4}
\end{align}
Furthermore, note that
\begin{align}
	&\frac{1}{\nworker}\sum_{i=1}^\nworker \vecnorm{\tg_i(\bx_k^{(i)})}^2 \\
	\leq& \frac{3}{\nworker}\sum_{i=1}^\nworker \vecnorm{\tg_i(\bx_k^{(i)}) -\tg_i(\by_k) }^2 + \nonumber \\
	    & \frac{3}{\nworker} \sum_{i=1}^\nworker\vecnorm{\tg_i(\by_k) -\tg(\by_k) }^2 + 3\vecnorm{\tg(\by_k) }^2  \\
	\leq& \frac{3\lip^2}{\nworker}\sum_{i=1}^\nworker\Exs\brackets{\vecnorm{\by_k-\bx_k^{(i)}}^2} + 3\kappa^2 + 3\vecnorm{\tg(\by_k) }^2. \label{eqn:deviation5}
\end{align}
Combing \Cref{eqn:deviation4,eqn:deviation5}, we have
\begin{align}
    &\parenth{1 - \frac{15\lr^2\lip^2\cp^2}{\glr^2}}\frac{1}{K\nworker}\sum_{k=0}^{K-1}\sum_{i=1}^\nworker\Exs\brackets{\vecnorm{\by_k-\bx_k^{(i)}}^2} \nonumber \\
    \leq& \lr^2\vbnd\brackets{\frac{2}{(2-\glr)\glr}\cp-1} + \frac{15\lr^2\cp^2\kappa^2}{\glr^2} + \nonumber \\
        & \frac{15\lr^2\cp^2}{\glr^2 K}\sum_{k=0}^{K-1}\Exs\brackets{\vecnorm{\tg(\by_k) }^2}.
\end{align}
For the ease of writing, define $D=15\lr^2\lip^2\cp^2/\glr^2$. Then,
\begin{align}
   &\frac{\lip^2}{K\nworker}\sum_{k=0}^{K-1}\sum_{i=1}^\nworker\Exs\brackets{\vecnorm{\by_k-\bx_k^{(i)}}^2} \nonumber \\
    \leq& \frac{\lr^2\lip^2\vbnd}{1-D}\brackets{\frac{2}{(2-\glr)\glr}\cp-1} + \frac{\lr^2\lip^2\cp^2\kappa^2}{\glr^2(1-D)} + \nonumber \\
        & \frac{D}{1-D} \frac{1}{K}\sum_{k=0}^{K-1}\Exs\brackets{\vecnorm{\tg(\by_k) }^2}. \label{eqn:res1}
\end{align}
Substituting \Cref{eqn:res1} into \Cref{eqn:res0}, one can get
\begin{align}
    &\frac{1}{K}\sum_{k=0}^{K-1}\Exs\brackets{\vecnorm{\tg(\by_k)}^2} \nonumber \\
    &\leq \frac{2[\obj(\by_0)-\obj_\text{inf}]}{\efflr K} + \frac{\efflr\lip\vbnd}{\nworker} + \nonumber \\
        & \frac{\lr^2\lip^2\vbnd}{1-D}\brackets{\frac{2}{(2-\glr)\glr}\cp-1} + \frac{\lr^2\lip^2\cp^2\kappa^2}{\glr^2(1-D)} + \nonumber \\
        & \frac{D}{1-D} \frac{1}{K}\sum_{k=0}^{K-1}\Exs\brackets{\vecnorm{\tg(\by_k) }^2}.
\end{align}
After minor rearranging, it follows that
\begin{align}
    &\frac{1}{K}\sum_{k=0}^{K-1}\Exs\brackets{\vecnorm{\tg(\by_k)}^2} \nonumber \\
    &\leq \brackets{\frac{2[\obj(\by_0)-\obj_\text{inf}]}{\efflr K} + \frac{\efflr\lip\vbnd}{\nworker}}\frac{1-D}{1-2D} + \nonumber \\
        & \frac{\lr^2\lip^2\vbnd}{1-2D}\brackets{\frac{2}{(2-\glr)\glr}\cp-1} + \frac{\lr^2\lip^2\cp^2\kappa^2}{\glr^2(1-2D)}.
\end{align}
When the learning rate is set to $\lr=\frac{1}{\lip}\sqrt{\frac{\nworker}{K}}$, $D = 15\nworker\cp^2/(\glr^2 K)$. If $K \geq 60\nworker\cp^2/\glr^2$, then $1-2D \geq 1/2$ and hence,
\begin{align}
    &\frac{1}{K}\sum_{k=0}^{K-1}\Exs\brackets{\vecnorm{\tg(\by_k)}^2} \nonumber \\
    \leq& \frac{4[\obj(\by_0)-\obj_\text{inf}]}{\efflr K} + \frac{2\efflr\lip\vbnd}{\nworker} + \nonumber \\
        & 2\lr^2\lip^2\vbnd\brackets{\frac{2}{(2-\glr)\glr}\cp-1} + \frac{2\lr^2\lip^2\cp^2\kappa^2}{\glr^2} \\
    =&  \frac{4\lip[\obj(\by_0)-\obj_\text{inf}]}{(1-\glr)\sqrt{\nworker K}} + \frac{2(1-\glr)\vbnd}{\sqrt{\nworker K}} + \nonumber \\
     & \frac{2\nworker\vbnd}{K}\brackets{\frac{2}{(2-\glr)\glr}\cp-1} + \frac{2\nworker\cp^2\kappa^2}{\glr^2 K} \\
    =& \mathcal{O}\parenth{\frac{1}{\sqrt{\nworker K}}} + \mathcal{O}\parenth{\frac{1}{K}}.
\end{align}
Here we complete the proof of \Cref{thm:main}.

\section{Additional Experimental Results}
\begin{figure}[!htb]
    \centering
    \begin{subfigure}{.24\textwidth}
    \centering
    \includegraphics[width=\textwidth]{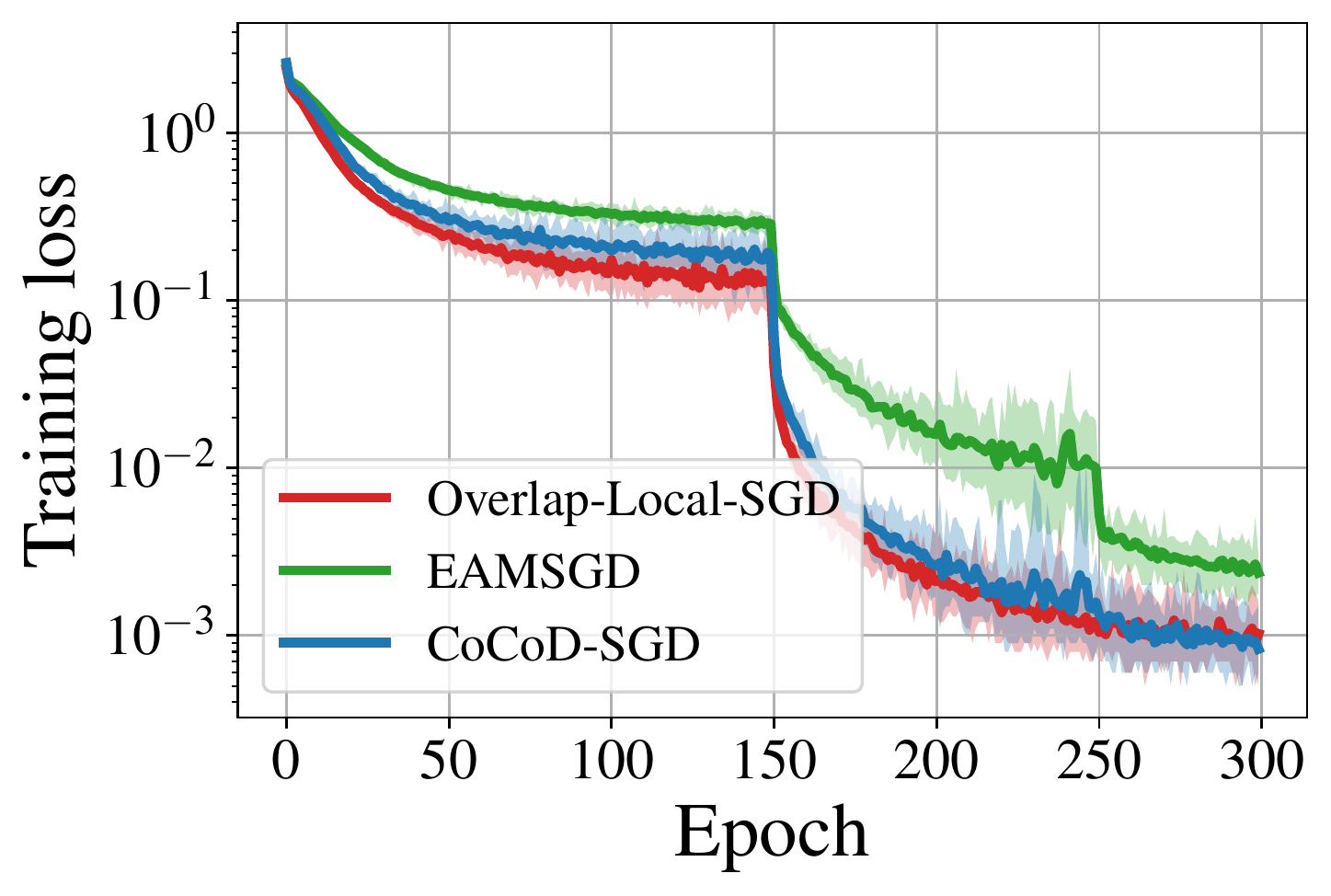}
    \end{subfigure}%
    ~
    \begin{subfigure}{.24\textwidth}
    \centering
    \includegraphics[width=\textwidth]{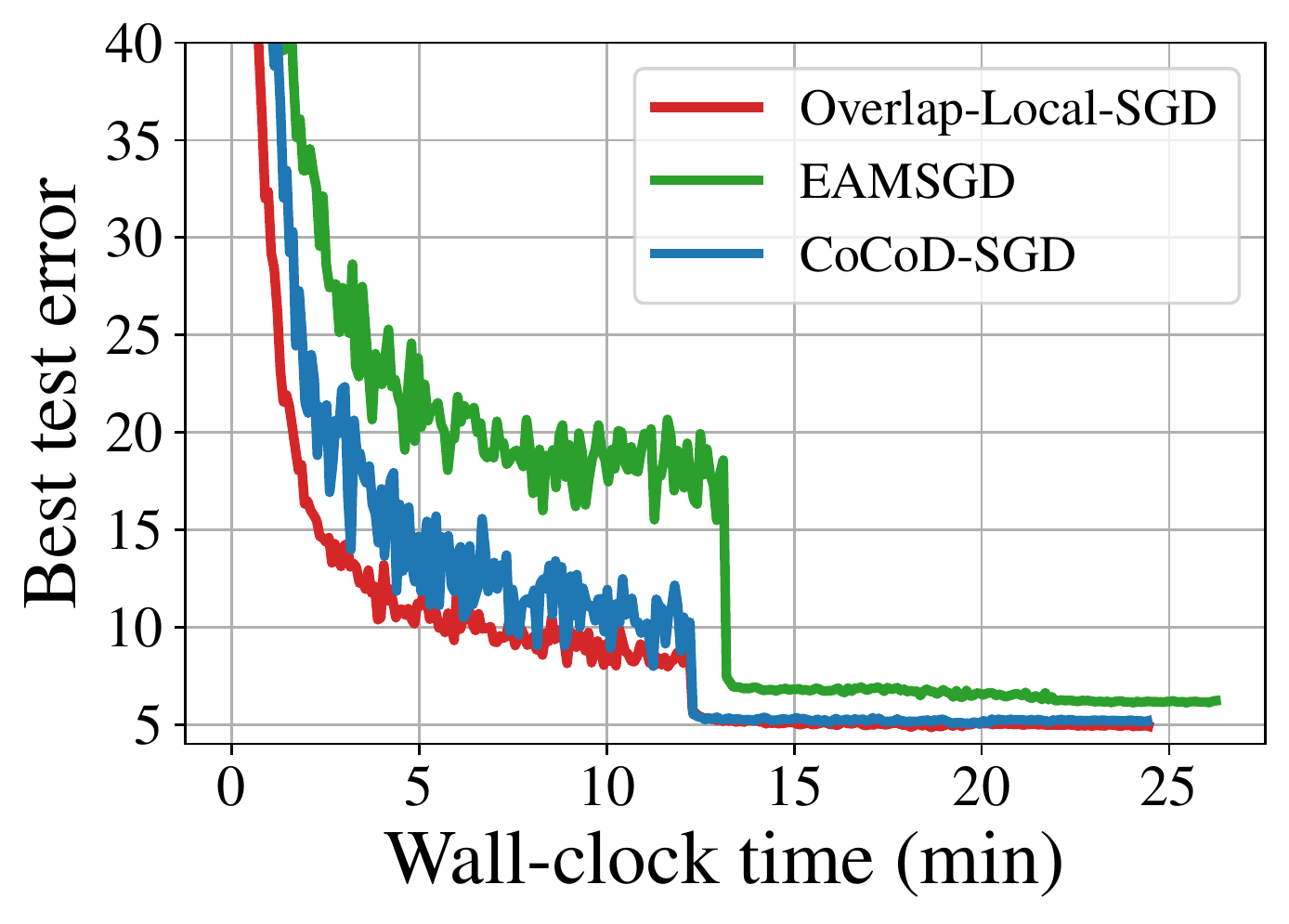}
    \end{subfigure}
    \caption{Comparison to CoCoD-SGD \cite{Shen2019FasterDD} and EAMSGD \cite{Zhang2015elasticsgd}. In all algorithms, the number of local updates $\cp$ is fixed as $2$. {\alg} slightly improves the loss-versus-iterations convergence of CoCoD-SGD.}
    \label{fig:cocod_curves}
\end{figure}

\end{document}